\documentclass[journal]{IEEEtran}
\usepackage{cite}
\usepackage{xcolor}
\usepackage{amsmath}
\usepackage{graphicx}
\usepackage[caption=false,font=footnotesize,labelfont=rm,textfont=rm]{subfig}
\usepackage[normalem]{ulem}
\graphicspath{{img/}{figures/}}

\begin{document}

\title{\vspace{-0.75cm}\huge Exploiting Adaptive Channel Pruning for Communication-Efficient Split Learning}

\author{Jialei Tan,
	Zheng Lin,
	Xiangming Cai,~\IEEEmembership{Senior Member,~IEEE,}
	Ruoxi Zhu,
	Zihan Fang, \\
	Pingping Chen,~\IEEEmembership{Senior Member,~IEEE},
	and Wei Ni,~\IEEEmembership{Fellow, IEEE } 

	\thanks{Jialei Tan, Xiangming Cai, and Pingping Chen are with the School of Physics and
	Information Engineering, Fuzhou University, Fuzhou 350108, China (e-mail: tanjlei@163.com; xiangming.cai@fzu.edu.cn; ppchen.xm@gmail.com).
	
	
	Zheng Lin is with the Department of Electrical and Electronic Engineering, The University of Hong Kong, Hong Kong. (e-mail: linzheng@eee.hku.hk).
	
	Ruoxi Zhu is with the School of Electrical Engineering and Telecommunications, University of New South Wales Sydney, Sydney, NSW 2052, Australia (e-mail: ruoxi.zhu@student.unsw.edu.au).

	Zihan Fang is with the Hong Kong JC STEM Lab of Smart City and the Department of Computer Science, City University of Hong Kong, Kowloon, Hong Kong SAR, China (e-mail: zihanfang3-c@my.cityu.edu.hk).
	
	Wei Ni is with Data61, CSIRO, Marsfield, NSW 2122, Australia, and the
	School of Computing Science and Engineering, and the University of New
	South Wales, Kennington, NSW 2052, Australia (e-mail: wei.ni@ieee.org).
	
    }
    \vspace{-0.8cm}

}

\maketitle

\begin{abstract}
	
Split learning (SL) transfers most of the training workload to the server, which alleviates computational burden on client devices. However, the transmission of intermediate feature representations, referred to as smashed data, incurs significant communication overhead, particularly when a large number of client devices are involved. To address this challenge, we propose an adaptive channel pruning-aided SL (ACP-SL) scheme. In ACP-SL, a label-aware channel importance scoring (LCIS) module is designed to generate channel importance scores, distinguishing important channels from less important ones. Based on these scores, an adaptive channel pruning (ACP) module is developed to prune less important channels, thereby compressing the corresponding smashed data and reducing the communication overhead. Experimental results show that ACP-SL consistently outperforms benchmark schemes in test accuracy. Furthermore, it reaches a target test accuracy in fewer training rounds, thereby reducing communication overhead.

\end{abstract}


\vspace{-0.0cm}

\section{Introduction}\label{section1}

With the proliferation of the Internet of Things (IoT) devices, an unprecedented amount of data is being generated. According to the International Data Corporation, the global data volume is projected to reach approximately 290~zettabytes by 2027~\cite{b1,lin2024efficient,fang2026hfedmoe,lin2024adaptsfl}. This unprecedented volume of data provides abundant support for machine learning (ML) algorithms~\cite{hong2026conflict,lin2025hsplitlora,peng2024sums,zhang2026transformer,lin2025hierarchical,fang2025dynamic,duan2025llm}. Conventional ML schemes typically adopt a centralized learning paradigm, in which massive raw data from client devices is transmitted to a central server for model training. However, this approach exposes raw data, leading to significant privacy leakage risks, and incurs substantial communication overhead, making practical deployment challenging~\cite{lin2024split}.

Federated learning (FL) has emerged to address the inherent limitations of centralized learning~\cite{b2,fang2024automated,lin2024fedsn}. In FL, client devices first train local models using their local datasets and then transmit the updated models to a server for model aggregation. As deep learning models continue to grow in size and complexity, the full-model on-device training paradigm in FL becomes increasingly impractical. For example, the recent on-device large model Mistral~7B presented in~\cite{b3} contains 7~billion parameters, making it challenging to train such models on resource-constrained edge devices via FL.

To overcome the limitations of FL, split learning (SL) partitions a model along the layer dimension into client-side and server-side components, thereby shifting most of the training workload to the server and reducing the computational burden on client devices \cite{b4,lin2025hasfl}. During training, the client and server exchange intermediate feature representations, referred to as smashed data. As the number of clients increases, the high communication overhead caused by smashed data transmission becomes a bottleneck for SL \cite{lin2024splitlora,b4-a,b4-b}.

Recent studies have proposed various smashed data compression techniques to address the above-mentioned issue. For instance, a SplitFedZip scheme was proposed in \cite{b5}, which employs an auto-encoder architecture to learn more compact intermediate representations. In \cite{b6}, a binarization SL scheme was proposed to uniformly compress client-side smashed data to 1~bit, thereby reducing communication overhead. Furthermore, a RandTopk SL scheme was proposed in \cite{b7}, which retains the top-$k$ elements with the largest magnitudes along with a small portion of non-top-$k$ components, enabling communication-efficient SL.

Although these schemes alleviate the communication overhead caused by large smashed data transmission, they typically apply uniform compression across all channels, overlooking the effect of the unequal importance of different channels on training performance. Specifically, some channels include task-relevant semantic information~\cite{b8}, whereas others are less informative or may even introduce noise. Ignoring these disparities can lead to suboptimal compression, potentially over-compressing smashed data associated with important channels while under-compressing smashed data corresponding to less important channels.

To address these challenges, we propose a communication-efficient adaptive channel pruning-aided SL (ACP-SL) scheme, aimed at reducing communication overhead without degrading its test accuracy. The proposed ACP-SL scheme comprises two modules: label-aware channel importance scoring (LCIS) and adaptive channel pruning (ACP). The LCIS module quantifies the importance scores of all channels, which are then used as a criterion for the ACP module to adaptively adjust the channel-wise pruning ratio at each iteration for channel pruning. Specifically, a larger channel importance score indicates that the channel is important and should be preserved by the ACP module. In contrast, a lower score indicates that the channel is less important, which is pruned by the ACP module.

\begin{figure}[t]
	\centering
	\vspace{-0.4cm}
	\includegraphics[width=0.48\textwidth]{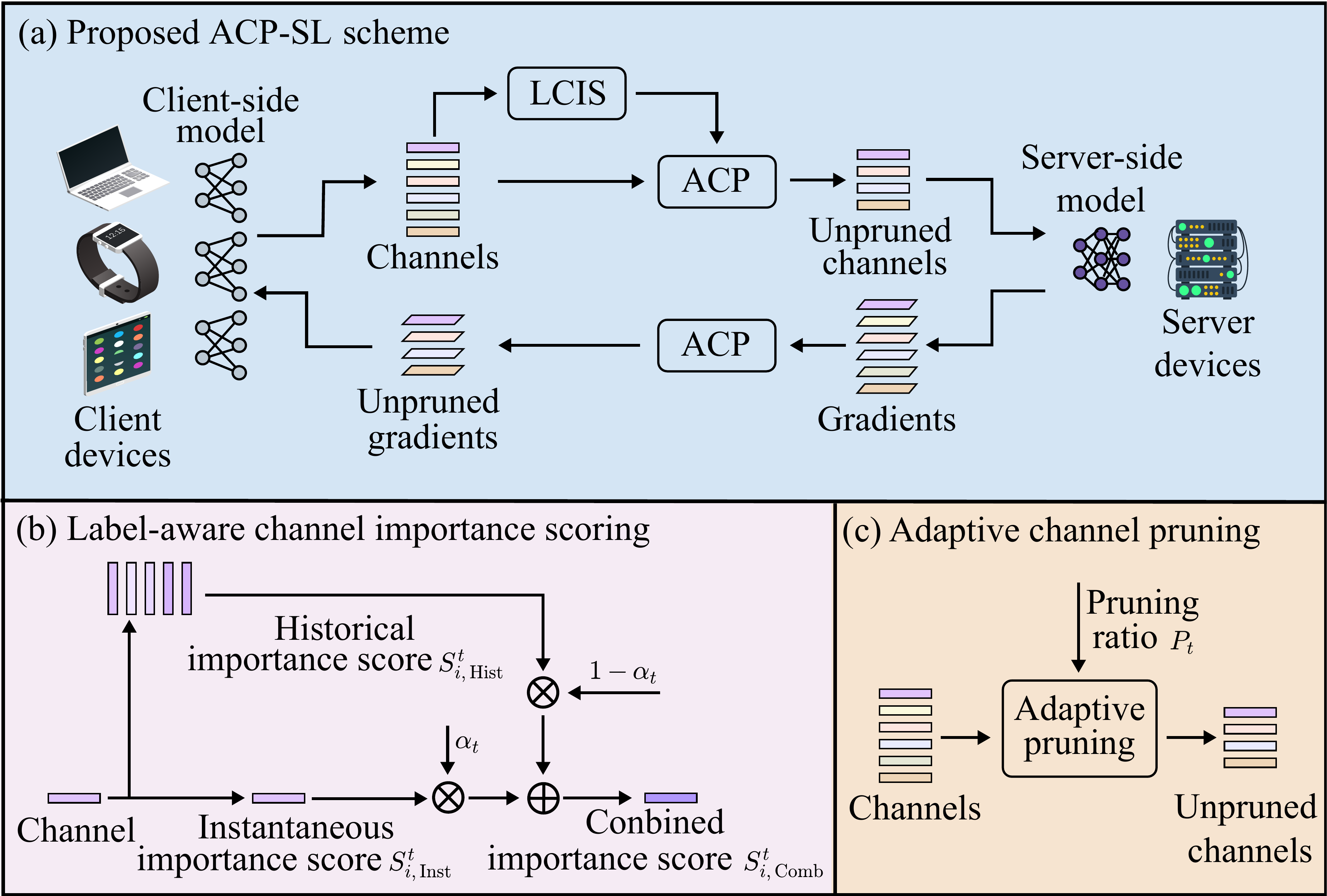}
	\caption{Block diagram of the proposed ACP-SL scheme.}\label{fig1}
	\vspace{-0.2cm}
\end{figure}

The contributions of this paper are summarized as follows:

\begin{enumerate}
	
\item We propose the LCIS module to quantify the importance score of each channel to model training. Using these scores, the LCIS module distinguishes important channels from less important ones.

\item We propose the ACP module to adaptively adjust the pruning ratio for channel pruning based on channel importance scores, thereby compressing smashed data. The ACP module preserves important channels essential for model training while pruning less important ones, reducing communication overhead caused by transmitting the smashed data associated with these less important channels.

\item Experimental results demonstrate that the proposed ACP-SL achieves higher test accuracy than the benchmark schemes. ACP-SL also requires fewer training rounds to reach a target test accuracy compared to the benchmarks, thereby reducing communication overhead.

\end{enumerate}

The remainder of this paper is organized as follows. Section~\ref{section2} presents the proposed ACP-SL scheme. Section~\ref{section4} provides experimental results and discussions. Finally, Section~\ref{section5} concludes the paper.

\section{Proposed ACP-SL scheme}\label{section2}

	Fig.~\ref{fig1} illustrates the block diagram of the proposed ACP-SL scheme. The global model is partitioned into client-side and server-side models, deployed on client devices and a server, respectively. Each ACP-SL iteration consists of six stages: (i) the client performs forward propagation of the client-side model on its local dataset to generate smashed data; (ii) the proposed LCIS module quantifies the importance score of each channel; (iii) the proposed ACP module employs an adaptive pruning ratio to prune channels based on these importance scores; (iv) the smashed data corresponding to unpruned channels is transmitted to the server, where forward and backward propagations of the server-side model are performed to compute gradients;
	(v) the ACP module employs the same pruning ratio to prune these gradients; and (vi) the client receives the unpruned gradients, performs local backpropagation, and updates the parameters of the client-side model.

	The proposed ACP-SL scheme comprises two modules: LCIS and ACP. Specifically, the LCIS module quantifies the importance score of each channel to model training, distinguishing important channels from less important ones, while the ACP module leverages these scores to adaptively adjust the channel pruning ratio. In the following, we elaborate on the principles of the proposed LCIS and ACP modules.

\begin{figure}[t]
	\centering
	\vspace{-0.45cm}
	\subfloat[Training loss of channel groups over a training round]{%
		\includegraphics[width=0.48\linewidth]{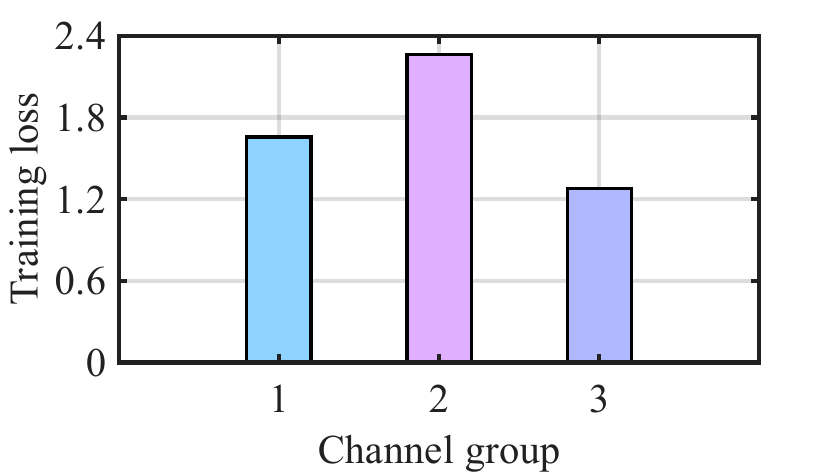}%
		\label{sim_fig1a}
	} \hspace{-0.03cm}
	\subfloat[Training loss of channel groups over multiple training rounds]{%
		\includegraphics[width=0.48\linewidth]{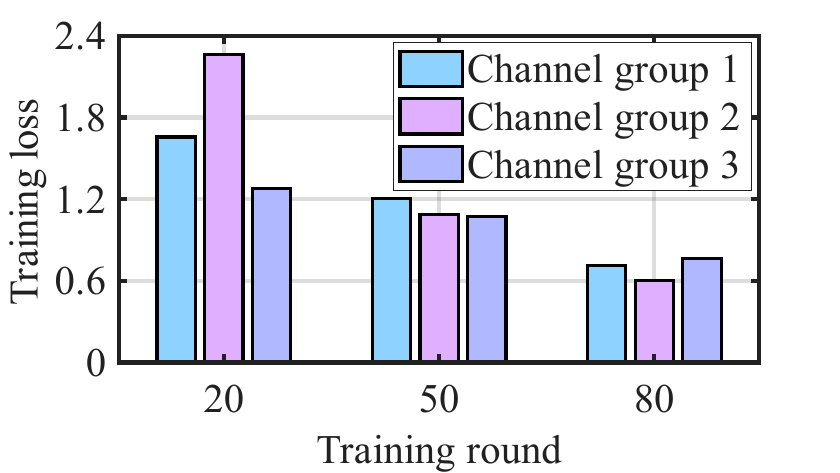}%
		\label{sim_fig1b}
	}
	\caption{Effect of channel importance on the training loss of SL.}
	\label{sim_fig1}
	\vspace{-0.4cm}
\end{figure}

\subsection{Label-Aware Channel Importance Scoring}\label{section3}

In conventional SL schemes, different channels generally contribute unequally to model training. After performing the channel group ablation experiments, Fig.~\ref{sim_fig1} illustrates the effect of channel importance on the training loss of SL, where ResNet-18 is adopted as the network model and MNIST is used as the dataset. As observed in Fig.~\ref{sim_fig1}, the importance across channel groups can be highly different, and the importance of each channel group varies across training rounds. Intuitively, communication overhead incurred by transmitting excessive smashed data corresponding to less important channels limits the practical deployment of conventional SL schemes. Consequently, pruning these less important channels in the smashed data can reduce transmission overhead and improve communication efficiency.

An important channel exhibits high intra-label similarity and lower inter-label similarity. In other words, samples with the same label are tightly clustered, while samples from different labels are substantially separated. Accordingly, the proposed LCIS module leverages both intra-label similarity and inter-label similarity to quantify the channel importance score, which is then used as a criterion for channel pruning in ACP. Specifically, the proposed LCIS module quantifies the channel importance score through three stages: 1) calculating instantaneous channel importance score, 2) calculating historical channel importance score, and 3) combining instantaneous and historical scores. These stages are elaborated as follows.

\subsubsection{Calculation of Instantaneous Channel Importance Score}

At iteration $t$, the instantaneous channel importance score consists of two components: the intra-label similarity score and the inter-label similarity score.

The intra-label similarity score measures the degree to which samples belonging to the same label are clustered within channel $i$. For a dataset with $K$ labels, the label-wise mean feature map corresponding to channel $i$ for label $n$, $n \in \{1, \dots, K\}$, is expressed as
\begin{equation}
    \bar{\mathbf{A}}_{i,n}^t = \frac{1}{|B_n|} \sum_{b \in B_n} \mathbf{A}_{i,b}^t,
\end{equation}
where $\mathbf{A}_{i,b}^t$ denotes the feature map of sample $b$ corresponding to channel $i$ at iteration $t$, $B_n$ is the set of samples with label $n$, and $|B_n|$ is the number of samples in $B_n$. Thus, the intra-label similarity score between the feature map $\mathbf{A}_{i,b}^t$ and the mean feature map $\bar{\mathbf{A}}_{i,n}^t$ is formulated as
\begin{equation}
    S_{i, \rm{Intra}}^t = \frac{1}{K} \sum_{n=1}^{K} \left( \frac{1}{|B_n|} \sum_{b \in B_n} \langle \mathbf{A}_{i,b}^t, \bar{\mathbf{A}}_{i,n}^t \rangle_F \right),
\end{equation}
where $\langle \cdot, \cdot \rangle_F$ denotes the Frobenius inner product. Finally, $S_{i,\rm{Intra}}^t$ is normalized across all $N$ channels to generate the normalized intra-label similarity score, denoted by $\bar{S}_{i,\rm{Intra}}^t$, which is given by
\begin{equation}
    \bar{S}_{i, \rm{Intra}}^t = \frac{S_{i, \rm{Intra}}^t}{\sum_{i=1}^{N} S_{i, \rm{Intra}}^t}.
\end{equation}

The inter-label similarity score measures the degree of similarity among different labels. It is defined as the average similarity score over all pairs of label-wise mean feature maps, formulated as
\begin{equation}
	S_{i, \rm{Inter}}^t = \frac{1}{\binom{K}{2}} \sum_{n=1}^{K-1} \sum_{m=n+1}^{K} \langle \bar{\mathbf{A}}_{i,n}^t, \bar{\mathbf{A}}_{i,m}^t \rangle_F,
\end{equation}
where $\bar{\mathbf{A}}_{i,n}^t$ and $\bar{\mathbf{A}}_{i,m}^t$ are the mean feature maps corresponding to label $n$ and label $m$, respectively; and $\binom{K}{2}$ denotes the number of label-wise mean feature maps pairs. Similarly, the normalized inter-label similarity score is obtained as
\begin{equation}
    {\bar{S}_{i, \rm{Inter}}^t} = \frac{S_{i, \rm{Inter}}^t}{\sum_{i=1}^{N} S_{i, \rm{Inter}}^t}.
\end{equation}

Consequently, the instantaneous importance score of channel $i$ at iteration $t$ is expressed as
\begin{equation}\label{eq6}
    S_{i,\rm{Inst}}^t = {\bar{S}_{i, \rm{Intra}}^t} - {\bar{S}_{i, \rm{Inter}}^t}.
\end{equation}
According to (\ref{eq6}), a larger intra-label similarity score ${\bar{S}_{i, \rm{Intra}}^t}$ and a lower inter-label similarity score ${\bar{S}_{i, \rm{Inter}}^t}$ result in a larger instantaneous channel importance score $S_{i,\rm{Inst}}^t$ for channel $i$.

\subsubsection{Calculation of Historical Channel Importance Score}

The instantaneous channel importance score $S_{i,\rm{Inst}}^t$ is sensitive to the instantaneous noise and outliers, which may lead to pruning important channels. Consequently, a historical channel importance score is introduced to alleviate the adverse effect of such incorrect pruning. The historical channel importance score ${S}_{i,\rm{Hist}}^t$ is quantified as the average of $t$ instantaneous channel importance scores, $\{ {S_{i,{\rm{Inst}}}^\tau } \}_{\tau = 1}^t$, and formulated as
\begin{equation}
    {S}_{i,\rm{Hist}}^t = \frac{1}{t} \sum_{\tau=1}^{t} S_{i,\rm{Inst}}^{\tau}.
\end{equation}

\begin{figure}[t]
	\centering
	\vspace{-0.3cm}
	\subfloat[Train loss vs. training round]{%
		\includegraphics[width=0.48\linewidth]{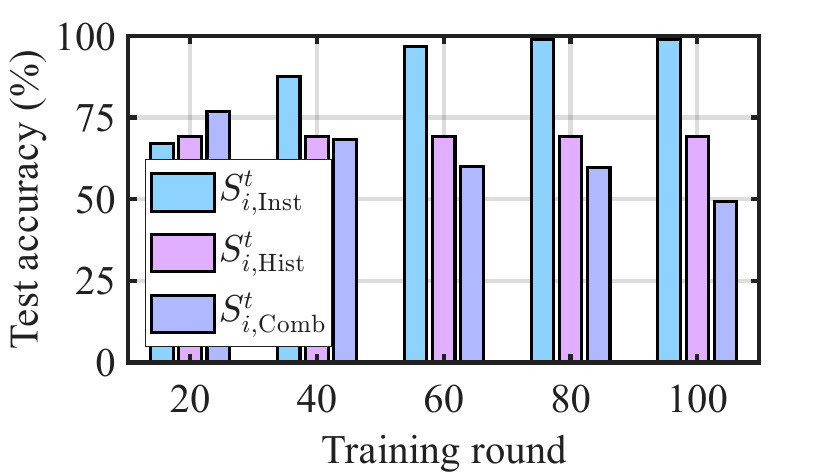}\label{sim_fig2a}
	}
	\subfloat[Test accuracy vs. training round]{%
		\includegraphics[width=0.48\linewidth]{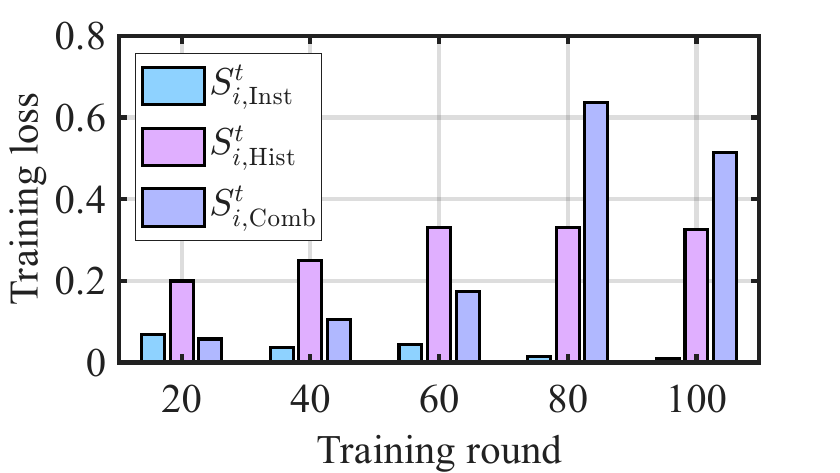}\label{sim_fig2b}
	}
	\caption{Effects of the instantaneous channel importance score $S_{i,\rm{Inst}}^t$, the historical channel importance score ${S}_{i,\rm{Hist}}^t$, and the combined channel importance score ${S}_{i,\rm{Comb}}^t$ on the training loss and test accuracy of SL.}
	\label{sim_fig2}
	\vspace{-0.0cm}
\end{figure}

\subsubsection{Combination of Instantaneous and Historical Scores}

When model training merely employs the instantaneous channel importance score $S_{i,\rm{Inst}}^t$, the test accuracy may improve rapidly during the initial phase. However, the accuracy subsequently degrades due to the the adverse effects of instantaneous noise and outliers, as $S_{i,\rm{Inst}}^t$ is sensitive to such perturbations. In contrast, the historical channel importance score ${S}_{i,\rm{Hist}}^t$, computed as the average of $\{ {S_{i,{\rm{Inst}}}^\tau } \}_{\tau = 1}^t$, is robust to noise and outliers. As a result, the instantaneous channel importance score $S_{i,\rm{Inst}}^t$ and the historical channel importance score ${S}_{i,\rm{Hist}}^t$ are combined to form the overall channel importance score, expressed as
\begin{equation}
    S_{i,\rm{Comb}}^{t} = \alpha_{t} S_{i,\rm{Inst}}^{t} + (1 - \alpha_{t}) {S}_{i,\rm{Hist}}^t,
\end{equation}
where $\alpha_t \in [0, 1]$ denotes the weighting coefficient. A large coefficient, $\alpha_t  \ge  \frac{1}{2} $, is applied during the initial phase of training, and a small coefficient, $\alpha_t  <  \frac{1}{2} $, is used in the later phase. The weighting coefficient is generated using a linearly decaying function, as given by
\begin{equation}
    \alpha_t= 1 - \frac{t}{T},
\end{equation}
where $t$ denotes the iteration index, and $T$ represents the total number of iterations. Subsequently, the resulting combined channel importance score $S_{i,\rm{Comb}}^{t}$ is fed into the ACP module for channel pruning.

Fig. \ref{sim_fig2} shows the effects of the instantaneous channel importance score $S_{i,\rm{Inst}}^t$, the historical channel importance score ${S}_{i,\rm{Hist}}^t$, and the combined channel importance score ${S}_{i,\rm{Comb}}^t$ on the training loss and test accuracy of SL. The pruning ratio is set to $0.99$. As observed in Fig. \ref{sim_fig2}, the training loss of the model using ${S}_{i,\rm{Comb}}^t$ consistently decreases compared to those using $S_{i,\rm{Inst}}^t$ and ${S}_{i,\rm{Hist}}^t$. Although the model using $S_{i,\rm{Inst}}^t$ achieves the highest test accuracy at the $20$th training round, its accuracy declines at the $40$th, $60$th, $80$th, and $100$th rounds. In contrast, the test accuracy of the model using ${S}_{i,\rm{Comb}}^t$ steadily increases, as the combined channel importance score integrates the initial gains of the instantaneous score with the robustness to noise and outliers provided by the historical score.

\subsection{Adaptive Channel Pruning}

Pruning less important channels to reduce the volume of smashed data is essential for achieving communication-efficient SL. However, most existing SL schemes employ static compression strategies for all smashed data, without considering the varying importance of different channels across iterations. This may result in over-compressing the smashed data corresponding to important channels while under-compressing the smashed data corresponding to less important ones. To address this limitation, we propose the ACP module, which utilizes the combined channel importance scores ${S}_{i,\rm{Comb}}^t$ to adaptively adjust the channel-wise pruning ratio and compress the smashed data accordingly.

At iteration $t$, the proposed ACP module computes the instantaneous channel group importance score by averaging the combined channel importance scores $S_{i,\rm{Comb}}^{t}$ across all $N$ channels, expressed as
\begin{equation}
    D_{\rm{Inst}}^t = \frac{1}{N} \sum_{i=1}^{N} S_{i,\rm{Comb}}^{t}.
\end{equation}
Subsequently, a historical channel group importance score is defined as the average of $t$ instantaneous channel group importance scores, $\{ {{ D_{\rm{Inst}}^\tau }} \}_{\tau  = 1}^t$, formulated as
\begin{equation}
	D_{\rm{Hist}}^{t} = \frac{1}{t} \sum_{\tau=1}^{t} D_{\rm{Inst}}^\tau.
\end{equation}

By dividing the historical channel group importance score $D_{\rm{Hist}}^t$ by the instantaneous channel group importance score $D_{\rm{Inst}}^{t}$, a scaling factor $W_{t}$ is obtained as follows:
\begin{equation}
	W_{t} = \frac{D_{\rm{Hist}}^{t}}{D_{\rm{Inst}}^t}.
\end{equation}
Notably, the scaling factor $W_{t}$ is used to adaptively adjust the channel-wise pruning ratio. Specifically, if $D_{\rm{Inst}}^t$ is larger than $D_{\rm{Hist}}^{t}$, then $W_t < 1$, resulting in a reduced pruning ratio to preserve important channels and their corresponding smashed data. Otherwise, the channels are considered less important and are pruned, and the corresponding smashed data is further compressed to reduce communication overhead.

To avoid sudden fluctuations in the pruning ratio, it is constrained within an interval of $[P_{\min}, P_{\max}]$, where $P_{\min}$ and $P_{\max}$ denote the minimum and maximum pruning ratios, respectively. Thus, the adaptive channel-wise pruning ratio at iteration $t$ in the proposed ACP module is formulated as \cite{b11}
\begin{equation}
	{P_t} = \max \left( {{P_{\min }},\min \left( {{W_t} \cdot {P_{{\rm{base}}}},{P_{\max }}} \right)} \right).
\end{equation}
where ${P_{{\rm{base}}}}$ denotes the pre-defined base pruning ratio. Notably, the ACP module utilizes the pruning ratio to adaptively prune less important channels, thereby compressing the smashed data compression and reducing communication overhead.

\section{Experimental Results and Discussions}\label{section4}

\subsection{Experimental Setup}

\textit{Implementation Details:} Experiments are conducted on an NVIDIA RTX 3090 GPU. To implement the SL environment on this single-GPU platform, clients are activated one at a time to simulate the training interactions between the client-side and server-side models, thereby ensuring a practical experimental setup.

\textit{Datasets and Models:} Both CIFAR-10 and Fashion-MNIST are used as datasets to evaluate the performance of the proposed ACP-SL scheme. Experiments are conducted under both independent and identically distributed (IID) and non-IID settings. Under the IID setting, samples are randomly shuffled and uniformly assigned to clients. Under the non-IID setting, heterogeneous client data partitions are generated using a Dirichlet distribution with parameter $\beta = 0.5$. The network model is ResNet-18, with the first four layers deployed on all clients as the client-side model and the remaining layers on the server as the server-side model.

\textit{Benchmarks:} To evaluate the effectiveness of the proposed ACP-SL, we benchmark it against three baseline schemes: Standard-SL, RandTopk-SL \cite{b7}, and Quantization-SL \cite{b12}.
\begin{itemize}
	\item \textbf{Standard-SL}: The standard SL scheme does not perform channel pruning and transmits the full smashed data without compression.
	
	\item \textbf{RandTopk-SL}~\cite{b7}: The RandTopk SL scheme transmits $k$ elements of smashed data, comprising primarily the largest-magnitude elements along with a small fraction randomly selected from the remaining elements.
	
	\item \textbf{Quantization-SL}~\cite{b12}: The quantization SL scheme applies quantization-based compression to smashed data.

\end{itemize}

\textit{Hyperparameters:} In the experiment, the learning rate of the stochastic gradient descent (SGD) is $0.00015$. The minimum pruning ratio $P_{\min}$ is $0.6$, the base pruning ratio $P_{\rm{base}}$ is $0.7$, and the maximum pruning ratio $P_{\max}$ is $0.8$. The mini-batch size is $128$, and the number of clients is $5$.

\begin{figure}[t]
	\centering
	\vspace{-0.5cm}
	\subfloat[\rmfamily\footnotesize CIFAR-10 (IID).]{%
		\includegraphics[width=0.48\linewidth]{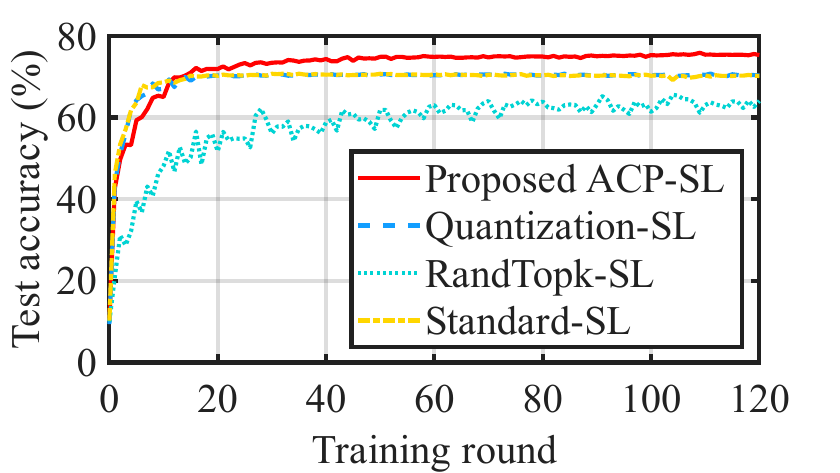}\label{sim_fig3a}
	}
	\subfloat[\rmfamily\footnotesize CIFAR-10 (non-IID).]{%
		\includegraphics[width=0.48\linewidth]{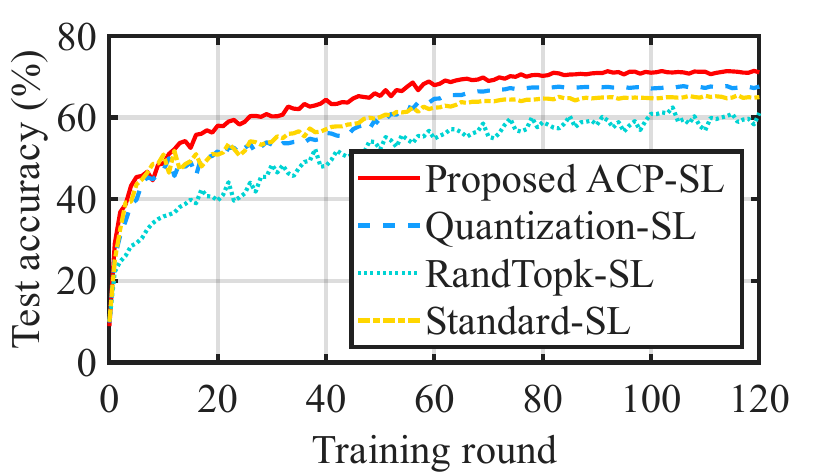}\label{sim_fig3b}
	}\\[-1.0ex]
	\subfloat[\rmfamily\footnotesize Fashion-MNIST (IID).]{%
		\includegraphics[width=0.48\linewidth]{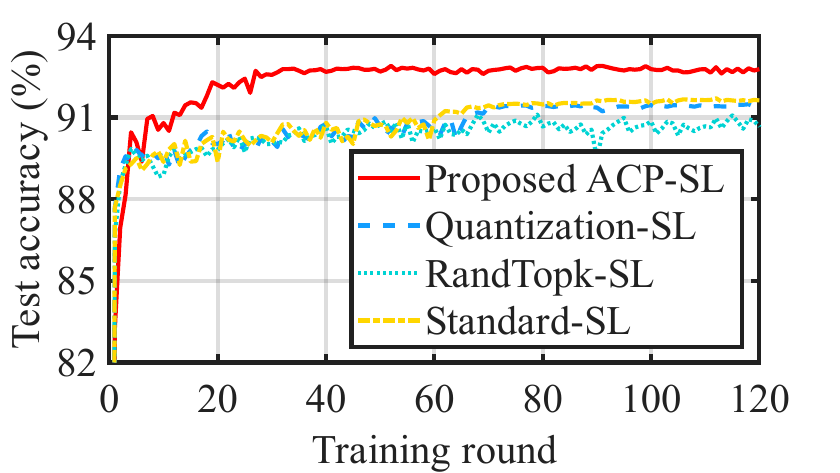}\label{sim_fig3c}
	}
	\subfloat[\rmfamily\footnotesize Fashion-MNIST (non-IID).]{%
		\includegraphics[width=0.48\linewidth]{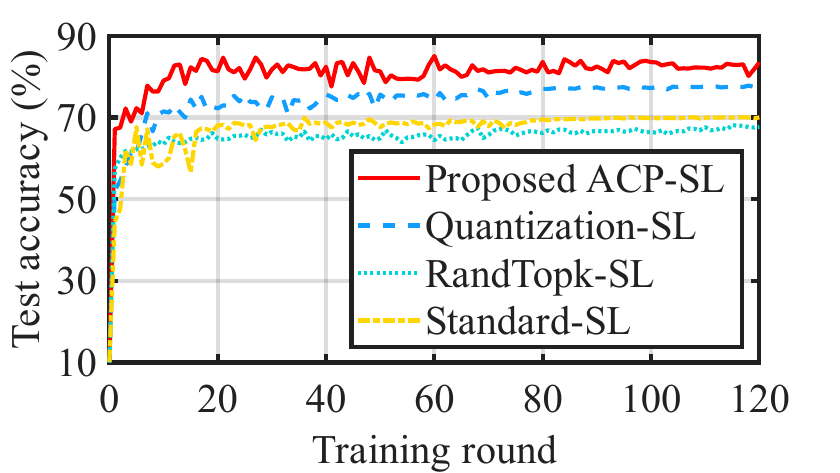}\label{sim_fig3d}
	}
	\caption{Test accuracy on CIFAR-10 and Fashion-MNIST under IID and non-IID settings.}
	\label{sim_fig3}
	\vspace{-0.2cm}
\end{figure}

\subsection{Comparison between ACP-SL and Benchmarks}

Fig.~\ref{sim_fig3} compares the proposed ACP-SL scheme with that of benchmarks under both IID and non-IID settings. The proposed ACP-SL scheme consistently achieves higher test accuracy than all benchmarks. Specifically, on the CIFAR-10 dataset, the ACP-SL scheme attains test accuracies of approximately $75.88$\% and $71.43$\%, surpassing Quantization-SL by about $5.11$\% and $3.72$\%, under IID and non-IID settings, respectively. On the Fashion-MNIST dataset, the ACP-SL scheme achieves test accuracies of approximately $92.90$\% and $85.09$\%, exceeding Quantization-SL by approximately $1.40$\% and $7.24$\%, under IID and non-IID settings, respectively.

The proposed ACP-SL scheme achieves the above-mentioned performance gain over the benchmark schemes primarily because it adopts the adaptive pruning ratio to prune less important channels, thereby compressing the corresponding smashed data. Meanwhile, important channels and their associated smashed data are preserved, which helps improve model performance, eventually leading to higher test accuracy. In contrast, Quantization-SL compresses smashed data indiscriminately without considering their relative importance. Such indiscriminate compression may degrade model performance and consequently limit achievable test accuracy.

The number of training rounds required to achieve a target test accuracy is defined as the communication overhead of an SL scheme. As illustrated in Figs.~\ref{sim_fig3a} and \ref{sim_fig3b}, on the CIFAR-10 dataset under the non-IID setting, the ACP-SL scheme requires about $46$ training rounds to reach a test accuracy of $65$\%, which is $12$ fewer rounds than that required by Quantization-SL. This result indicates that the proposed ACP-SL incurs lower communication overhead than Quantization-SL on the CIFAR-10 dataset. Similarly, as shown in Figs.~\ref{sim_fig3c} and \ref{sim_fig3d}, on the Fashion-MNIST dataset, ACP-SL requires fewer training rounds than Quantization-SL to achieve test accuracies of $91$\% and $75$\% under the IID and non-IID settings.

\begin{figure}[t]
	\centering
	\vspace{-0.5cm}
	\subfloat[\rmfamily\footnotesize CIFAR-10.]{%
		\includegraphics[width=0.48\linewidth]{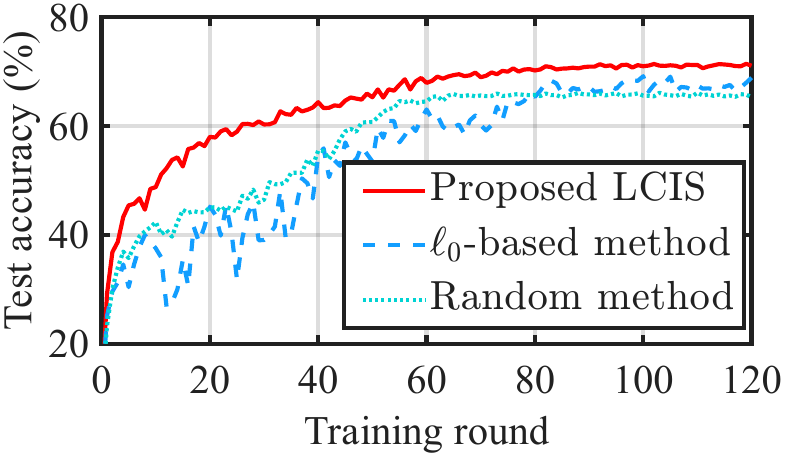}%
	}
	\subfloat[\rmfamily\footnotesize Fashion-MNIST.]{%
		\includegraphics[width=0.48\linewidth]{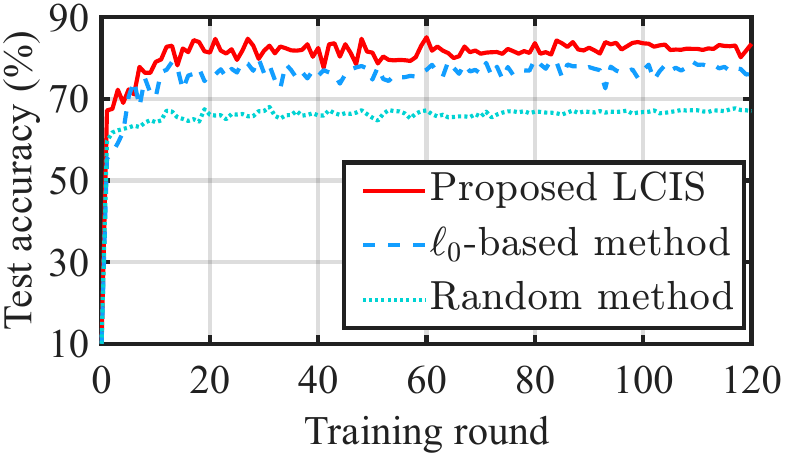}%
	}
	\caption{Ablation experiment results of the proposed LCIS on CIFAR-10 and Fashion-MNIST under the non-IID setting.}
	\label{sim_fig4}
\end{figure}

\begin{figure}[t]
	\centering
	\vspace{-0.5cm}
	\subfloat[\rmfamily\footnotesize CIFAR-10.]{%
		\includegraphics[width=0.48\linewidth]{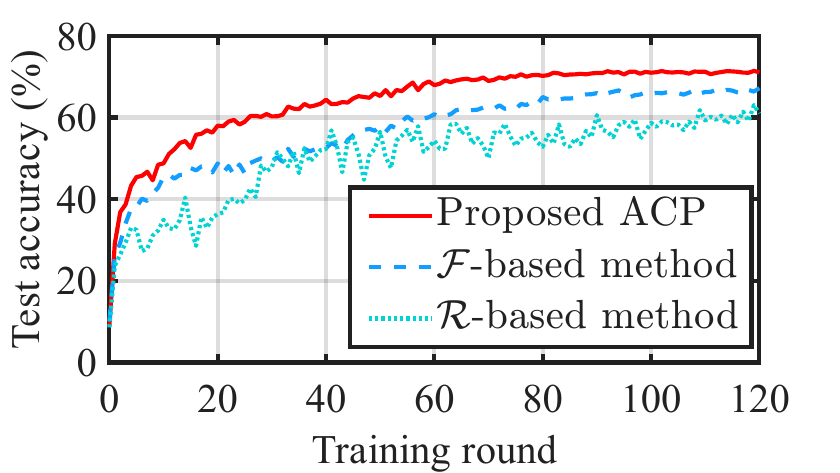}%
	}
	\subfloat[\rmfamily\footnotesize Fashion-MNIST.]{%
		\includegraphics[width=0.48\linewidth]{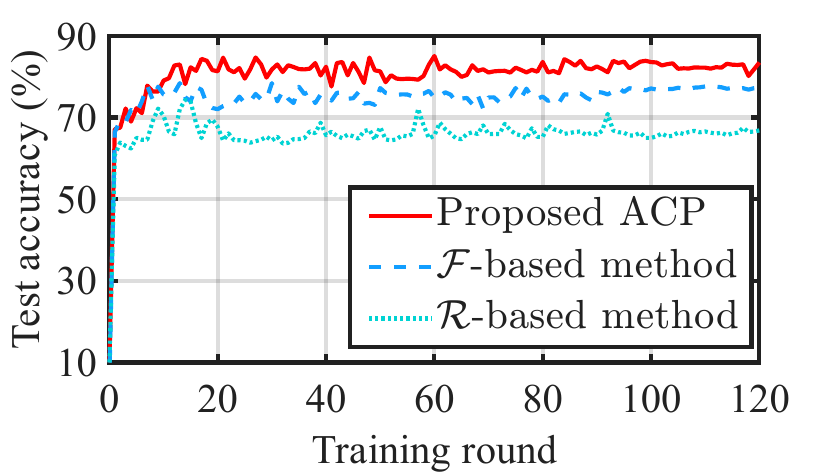}%
	}
	\caption{Ablation experiment results of the proposed ACP on CIFAR-10 and Fashion-MNIST under the non-IID setting.}
	\label{sim_fig5}
	\vspace{-0.3cm}
\end{figure}

\subsection{Ablation Study}

\subsubsection{Effect of LCIS on the Performance of ACP-SL}

Fig.~\ref{sim_fig4} examines the effect of the proposed LCIS on the test accuracy of the proposed ACP-SL scheme. In the $\ell_0$-based method, each channel is assigned to a channel importance score based on the number of non-zero elements in its corresponding feature map. Specifically, the channels corresponding to the feature maps containing more non-zero elements are considered more important and receive higher scores, while those corresponding to the feature maps with fewer non-zero elements are regarded as less important and receive lower scores. In the random method, a channel importance score is randomly assigned to each channel.

In Fig.~\ref{sim_fig4}, it is observed that the proposed ACP-SL scheme using LCIS achieves higher test accuracy than both $\ell_0$-based and random methods. The reason is that the random method does not differentiate between important and unimportant channels, and the feature map used in the $\ell_0$-based method is less effective than the label-wise feature map employed in the proposed LCIS.

\subsubsection{Effect of ACP on the Performance of ACP-SL}

Fig.~\ref{sim_fig5} presents the effect of the proposed ACP on the test accuracy of the proposed ACP-SL scheme. The ${\cal F}$-based method uses a fixed pruning ratio for channel pruning, while the ${\cal R}$-based method employs a random pruning ratio. As shown in Fig.~\ref{sim_fig5}, the ACP-SL scheme using the proposed ACP demonstrates higher test accuracy than the ${\cal F}$-based and ${\cal R}$-based methods. This is attributed to the fact that the proposed ACP combines both the instantaneous and historical channel group importance scores to generate an adaptive pruning ratio for channel pruning, whereas the benchmark methods rely solely on fixed or random pruning ratios, which degrades model performance and decreases test accuracy.

\section{Conclusions}\label{section5}

In this paper, we proposed the ACP-SL scheme to reduce the communication overhead incurred by transmitting smashed data. In ACP-SL, the LCIS module was introduced to evaluate the importance of each channel to model training and assign corresponding importance scores. Based on these scores, the ACP module adaptively prunes less important channels, thereby reducing the volume of transmitted smashed data and lowering communication overhead. Experimental results showed that ACP-SL achieves at least a $3.72$\% improvement in test accuracy on the CIFAR-10 dataset compared to the benchmark schemes. Additionally, ACP-SL reaches a test accuracy of $65$\% in approximately $46$ training rounds, reducing the required number of rounds by $12$ compared to Quantization-SL, to achieve the same accuracy. These results indicate that the proposed ACP-SL effectively reduces communication overhead while maintaining superior model performance.

\bibliographystyle{IEEEtran}
\bibliography{ref}

\end{document}